\documentclass[times, review, 10pt]{elsarticle}
\usepackage[numbers]{natbib}
\usepackage{amsmath} 
\usepackage{booktabs} 
\usepackage{graphicx}
\usepackage[export]{adjustbox}
\usepackage{multirow}
\usepackage{xspace}
\usepackage{url}

\usepackage{amssymb}    
\def\tsc#1{\csdef{#1}{\textsc{\lowercase{#1}}\xspace}}
\tsc{WGM}
\tsc{QE}
\tsc{EP}
\tsc{PMS}
\tsc{BEC}
\tsc{DE}

\begin{document}
\let\WriteBookmarks\relax
\def\floatpagepagefraction{1}
\def\textpagefraction{.001}

\title{UnGAP: Uncertainty-Guided Affine Prompting for Real-Time Crack Segmentation\tnoteref{t1}}

\tnotetext[t1]{This paper is supported in part by the Shanghai ``The Belt and Road'' Young Scholar Exchange Grant (24510742000).}

\author[monashit]{Conghui Li\corref{cor1}}
\ead{Conghui.Li@monash.edu}

\author[sjtu]{Huanyu He}
\ead{hehuanyu\_ee@sjtu.edu.cn}

\author[monasheng]{Xin Wang}
\ead{wang.xin@monash.edu}

\author[sjtu]{Weiyao Lin}
\ead{wylin@sjtu.edu.cn}

\author[monashit]{Chern Hong Lim\corref{cor1}}
\ead{lim.chernhong@monash.edu}

\affiliation[monashit]{organization={School of Information Technology, Monash University},
            city={Selangor},
            country={Malaysia}}

\affiliation[monasheng]{organization={School of Engineering, Monash University},
            city={Selangor},
            country={Malaysia}}

\affiliation[sjtu]{organization={Department of Electrical Engineering, Shanghai Jiao Tong University},
            city={Shanghai},
            country={China}}

\cortext[cor1]{Corresponding authors.}

\begin{abstract}
Real-time crack segmentation is vital for structural health monitoring but is plagued by aleatoric uncertainties arising from varying lighting, blur, and texture ambiguity. Current uncertainty-aware approaches typically treat uncertainty estimation as a passive endpoint for post-hoc analysis, failing to “close the loop” by feeding this information back to refine feature representations. We contend that independent pixel-wise heteroscedastic modeling is uniquely suited for crack segmentation, as cracks are defined by fine-grained local gradients rather than the global semantic coherence relied upon in general object segmentation. However, this approach suffers from a structural optimization pathology: high predicted variance attenuates loss gradients, effectively causing the model to ignore difficult samples and under-fit complex boundaries. To address these challenges, we propose UnGAP, a novel framework that establishes a closed-loop mechanism between uncertainty estimation and feature learning. Central to our approach is the Uncertainty-Prompted Feature Modulator (UPFM), which treats aleatoric uncertainty as an active visual prompt rather than a mere output. UPFM dynamically calibrates feature distributions through pixel-wise affine transformations. Crucially, this mechanism mitigates the heteroscedastic pathology by transforming high variance, which would otherwise indicate gradient suppression, into a constructive signal for stronger feature rectification in ambiguous regions. Additionally, a boundary-aware detection head is introduced to further constrain prediction precision. Extensive experiments demonstrate that UnGAP balances superior segmentation accuracy with real-time inference speed, effectively validating the benefit of transforming uncertainty from a passive metric into an active calibration tool.
\end{abstract}




\maketitle

\section{Introduction}

Crack segmentation plays a critical role across various fields, such as structural health monitoring. In infrastructure, real-time crack detection is vital to prevent catastrophic failures and to ensure the safety and longevity of buildings, bridges, and other structures; for example, Tingbin et al.~\cite{Tingbin2024} emphasized its importance in infrastructure inspection, while Shengyou et al.~\cite{Shengyou2024} further demonstrated its practical significance for structural safety. Tiny cracks can propagate over time, leading to significant deterioration and eventual collapse~\cite{Khan2024}. However, models often struggle with various sources of uncertainty, such as varying lighting conditions, occlusions, and inherent ambiguities in the visual appearance of cracks. These uncertainties can significantly complicate the segmentation process and decrease the accuracy of models. As such, a critical question is how to decrease the uncertainty or the influence of uncertainty at the segmentation model~\cite{Armen2009}. 



Faber classified uncertainty into aleatoric and epistemic types, where aleatoric uncertainty arises from data variability, and epistemic uncertainty stems from the model's limitations~\cite{Faber2005}. Building on this theory, uncertainty-aware segmentation has gained increasing attention. For example, Bayesian neural networks (BNNs) estimate predictive uncertainty by learning a distribution over model parameters rather than relying on deterministic values. Andrew et al.~\cite{Andrew2019} analyzed this perspective from Bayesian deep learning, while Audrey et al.~\cite{Audrey2021} further demonstrated its value for robust segmentation. Additionally, Monte Carlo dropout has been widely used to approximate epistemic uncertainty, improving prediction reliability~\cite{CHEN2021354}. Conversely, researchers acknowledge the inherent unpredictability of aleatoric uncertainty. To address this, stochastic segmentation methods are often used to output multiple possible outcomes by sampling multiple times, balancing accuracy with uncertainty calibration~\cite{ Lin2022, Xuming2022}. Another solution is to conduct pixel-wise uncertainty modeling through heteroscedastic regression~\cite{Kendall2017}.



\begin{figure*}
	\centering
\includegraphics[width=12 cm]{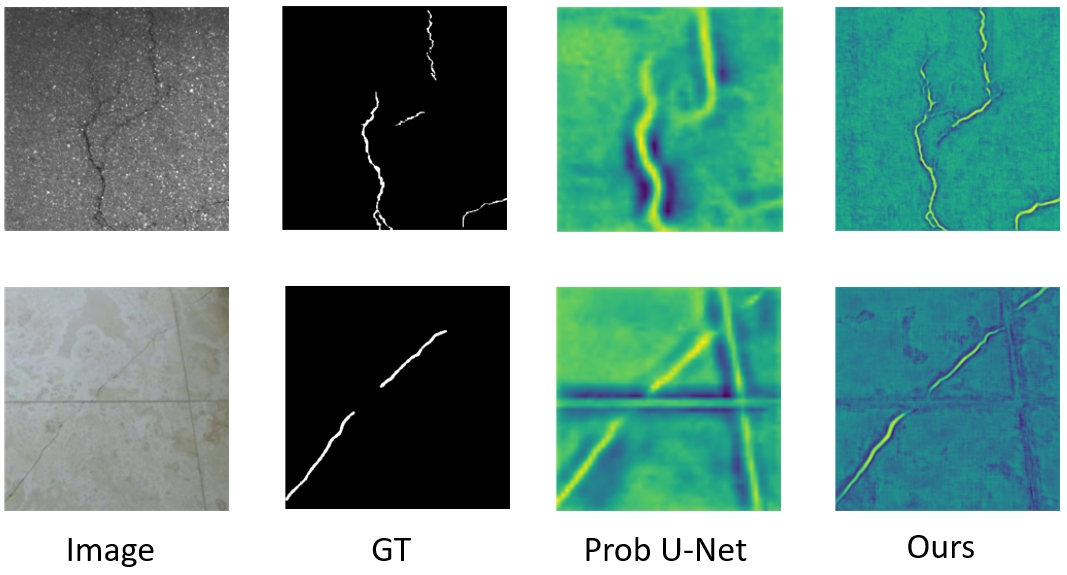}
\caption{Comparison of predicted heat map between UnGAP (Ours) and stochastic segmentation methods (Prob U-Net). Illustrating that the stochastic segmentation will cause blurred predictions, especially in complex environment. Instead, UnGAP shows distinct difference between object and background. GT represents Ground-truth.}\label{bayes}
\end{figure*}

However, a primary limitation of existing approaches is that they isolate uncertainty estimation from the feature learning process. These methods typically generate uncertainty maps solely for post-hoc analysis, visualization, or safety monitoring, without integrating this information back into the network to enhance performance. Consequently, the estimated uncertainty contributes neither to the refinement of feature representations nor to the improvement of segmentation accuracy. Furthermore, stochastic segmentation methods that introduce randomness via tiling operations often result in low-confidence, blurred outputs (as observed in Fig.~\ref{bayes}), which compromises the precision required for delineating fine crack boundaries.

More critically, while heteroscedastic regression is a promising direction, recent studies~\cite{Seitzer2022, WongToi2023} have identified a fundamental optimization issue in this approach: the loss gradient is weighted by the inverse of the predicted variance ($1/\sigma^2$). As a result, the model tends to predict high variance for challenging samples (e.g., those with complex textures) to minimize the loss, thereby attenuating the gradients in regions that require robust feature extraction. In dense prediction tasks, this leads to an imbalance where the network under-fits complex boundaries by assigning high variance instead of learning discriminative representations.

\begin{figure}
	\centering
\includegraphics[width=12 cm]{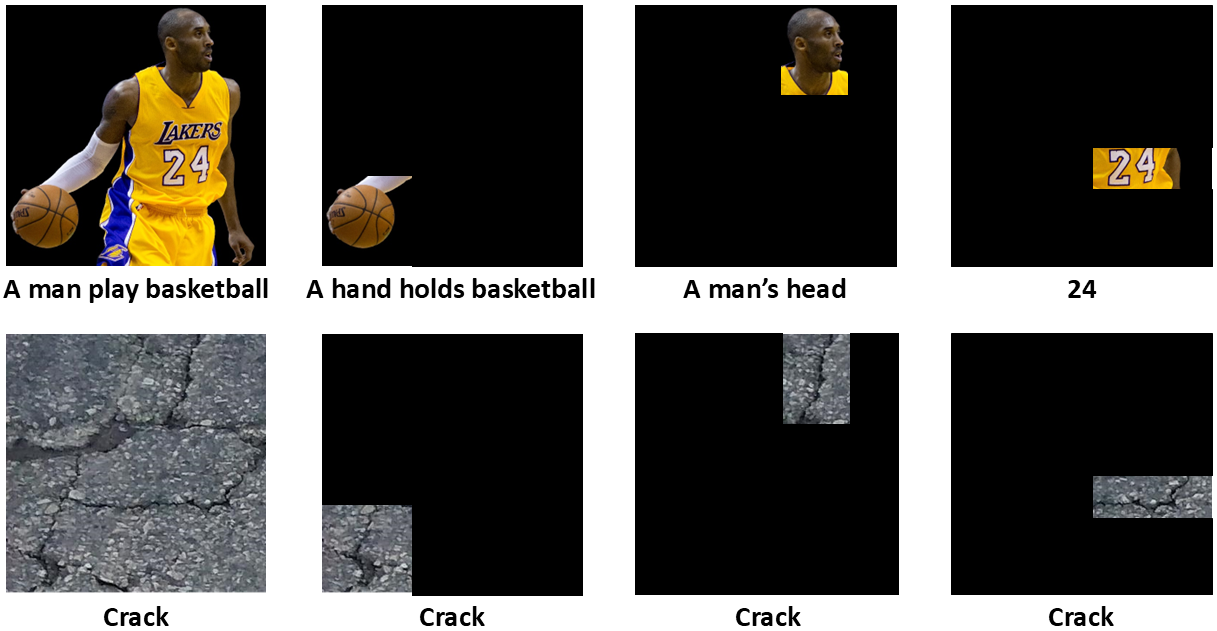}
\caption{Contrasting the importance of global context vs. local features. For conventional object recognition (top row), global context is crucial; removing parts of the image destroys key semantic information. For crack segmentation (bottom row), the defining characteristics are fine-grained local textures and gradients. The essential information for segmentation is preserved even in small crops. This motivates our focus on resolving local uncertainty rather than capturing global context.}\label{long_distance}
\end{figure}

Meanwhile, in general semantic segmentation, independent pixel-wise uncertainty is often criticized for failing to reflect object-level semantic reliability, as these tasks rely heavily on the coherence of large, structured objects (one example can be referred at top row of Fig.~\ref{long_distance}). For instance, vision transformers~\cite{Dosovitskiy2020} leverage global context and long-range dependencies to obtain the semantic understanding and structured objects. Due to these shortcomings, pixel-wise uncertainty estimation method has not been used widely. However, we argue that crack segmentation presents a fundamental exception where pixel-wise heteroscedastic modeling is conceptually aligned with the problem's physical nature. Unlike generic objects, cracks are primarily driven by fine, irregular textures and sharp boundary gradients rather than global shape or trajectory. Stochastic noise factors, such as blur, lighting variations, and surface contamination, vary significantly at specific spatial locations. From this perspective, independent pixel-level uncertainty is not a limitation but a necessity whereas regions surrounded by tiny, low-contrast cracks naturally exhibit higher data uncertainty than clean backgrounds. Consequently, relying on global semantics to rescue local ambiguities is ineffective, as the global shape of a crack provides little discriminative information compared to local visual features (see the bottom row of Fig.~\ref{long_distance}). Therefore, we hypothesize that the optimal strategy is to retain this physically consistent uncertainty modeling but explicitly resolve its optimization pathology. This implies that we must enable the model to locally calibrate itself using the uncertainty information, transforming high variance from a gradient suppressor into a feature refinement cue.

To address these challenges, we propose a novel framework named UnGAP. Unlike prior approaches that treat uncertainty estimation as an isolated output, UnGAP explicitly integrates the estimated heteroscedastic map back into the feature learning process. We propose an Uncertainty-Prompted Feature Modulator (UPFM) that utilizes aleatoric uncertainty as a conditioning input to spatially modulate the feature maps. Specifically, UPFM predicts pixel-wise affine transformation parameters to adaptively calibrate the feature distribution. In high-uncertainty regions, where standard heteroscedastic regression tends to attenuate gradients, UPFM leverages the variance signal to apply explicit feature scaling and shifting. This operation enhances the discriminability of ambiguous features by repositioning them away from the decision boundary, effectively transforming the high variance from a source of optimization instability into a guide for feature refinement. Additionally, given the irregular topology of cracks, a boundary-aware detection head is introduced to enforce geometric constraints and reduce false positives. Our lightweight design balances segmentation accuracy with inference speed, making it suitable for real-time applications.


The contributions of this work are as follows:
\begin{enumerate}

\item We revisit and justify pixel-wise heteroscedastic modeling for crack segmentation, arguing that cracks are dominated by local textures, gradients and spatially varying noise, making independent pixel-level uncertainty conceptually aligned with this task. Building on this insight, we adopt a Heteroscedastic Module (HM) trained with the $\beta$-NLL objective to produce reliable uncertainty maps as functional signals rather than passive outputs. 

\item We propose an Uncertainty-Prompted Feature Modulator (UPFM) to establish a closed-loop mechanism that feeds the predicted uncertainty map back into the decoder’s intermediate feature maps (i.e., the feature stream before the final prediction) via pixel-wise affine modulation. This converts the typically dampening high-variance signal from being a loss/gradient suppressor in heteroscedastic regression into a constructive cue for feature rectification in ambiguous crack regions.

\item Motivated by the boundary-centric feature and irregular topology of cracks, we integrate a lightweight Boundary-aware Detection Head with an explicit boundary branch to enhance gradient cues and sharpen crack delineation. The overall design achieves a favorable accuracy–speed balance for real-time crack segmentation.

\end{enumerate}

\section{Related work}

\subsection{Real-time segmentation}

In the literature, real-time segmentation can be categorized into two architectures: Dilated CNN Architecture and Symmetric Network Architecture. Dilated CNN Architecture allows the network to capture multi-scale contextual information without increasing the number of parameters or reducing the spatial resolution of the feature maps. Typically, DeepLabv3 incorporates atrous convolution to capture multi-scale contextual information, this allow model to adjust the receptive field~\cite{Liang2017}. The DeepCrack utilizes deep hierarchical convolutional layers to refine the crack prediction map, ensuring to maintain information from different receptive fields~\cite{Yahui2019}. 

On the other hand, Symmetric Network Architecture exhibits a balanced, mirror-like structure, which is able to maintain spatial resolution and preserve detailed features while encoding and decoding information. Unet introduced a symmetric encoder-decoder structure with skip connections to preserve the essential boundary information and small-scale features~\cite{Ronneberger2015}. HRNet continuously exchanges information across parallel layers that maintain different resolutions, leading to superior segmentation performance in complex tasks~\cite{Wang2021}. However, due to high inference cost of maintaining different resolutions and information exchange across different resolutions, Yongshang et al. proposed HrSegNet by maintaining only one low resolution at one stage to improve the efficiency~\cite{Yongshang2023}.

The logic of attention mechanism is aligned with our proposed method. Attention mechanism is a way to dynamically determining the importance of different parts of the input data~\cite{Dosovitskiy2020}. Therefore, beyond these two architectures, many studies have developed attention mechanisms for crack segmentation, inspired by the success of ViT. For example, Crackformer combines the strengths of CNNs for local feature extraction with Transformers' ability to capture global context and long-distance dependencies to improve the accuracy in crack task~\cite{Huajun2021}. PSANet introduces an innovative point-wise spatial attention mechanism that enables the network to aggregate contextual information more effectively than traditional methods to capture relationships between distant pixels~\cite{Zhao2018}. Recently, Mamba framework as a different multi-scale attention is used in crack segmentation for better spatial and long-range dependencies~\cite{Jianming2025}. Moreover, CCDFormer~\cite{Xiangkun2025} proposed a simplified transformer-based model to enhance feature extraction with dual-backbone U-shaped structure.

\subsection{Uncertainty}

Two primary types of uncertainty are recognized by Faber~\cite{Faber2005}: Aleatoric and Epistemic. Aleatoric uncertainty, also known as statistical or inherent uncertainty, arises from the inherent variability or randomness in the data or environment. This type of uncertainty is often irreducible because it is associated with the natural stochasticity present in the processes being modeled~\cite{Khanzhina2023}. For example, crack images with poor lighting, occlusions, superfine or super tiny objects might lead to high aleatoric uncertainty. Aleatoric uncertainty in this context is irreducible because it is tied to the data itself—no matter how well we refine the model, the inherent noise in the image remains. On the other hand, epistemic uncertainty, also known as systematic or model uncertainty, arises from a lack of knowledge or incomplete information about the model itself. This type of uncertainty is reducible because it stems from limitations in the model or insufficient data. Epistemic uncertainty indicates how confident the model is in its predictions; it tends to be higher when the model encounters new, unseen data or when the training data is sparse or unrepresentative~\cite{Eyke2021}. Unlike Aleatory uncertainty, Epistemic uncertainty can be reduced by improving the model. 


\subsection{Uncertainty-aware segmentation}

Early uncertainty-aware methods like Bayesian SegNet use dropout to approximate Bayesian inference but only provide pixel-wise probability maps~\cite{Alex2015}. More advanced methods, like Probabilistic U-Net and PHiSeg, use Variational Autoencoders (VAEs) to model the entire segmentation map's likelihood~\cite{Kohl2018,Baumgartner2019}. Inspired with these methods for general task, some works in crack segmentation that use simple VAE or Bayesian neural networks for aleatoric uncertainty estimation~\cite{Ketson2025, Rahul2023}. However, in these approaches, the posterior distribution of the VAE is often overly simplistic, making it difficult to model the true image distribution. As a result, the sampled outputs may lack meaningful semantic information. Although some studies have attempted to improve this aspect~\cite{Selvan2020,Zhang20222}, in real-time crack segmentation tasks, directly applying the latent variables which are sampled from the posterior distribution to feature maps through a ``tile operation" ``tile operation" may lead to blurred predictions, especially for small objects. Furthermore, the posterior distribution does not explicitly capture the intrinsic characteristics between image pixels or the noise, which make their results of uncertainty quantification less convincing. Lastly, such quantification results are often applied only as a reference but difficult to leverage directly to enhance model's segmentation performance. In contract, injecting randomness into segmentation model can even lead to a drop in performance. Unlike these methods, which inject randomness to generate multiple outputs, our model directly integrates aleatoric uncertainty into the feature map. This allows the model to focus on challenging areas without the overhead of multiple inferences. 



To address these issues, we adopt a heteroscedastic aleatoric formulation that estimates pixel-wise variance (log-variance) instead of producing multiple stochastic outputs. Motivated by the boundary-centric and texture-driven nature of cracks, we focus on heteroscedastic aleatoric modeling and further leverage the resulting uncertainty map as an explicit conditioning signal for feature refinement in a closed-loop design: the predicted variance highlights ambiguous regions and is fed back into the decoder feature stream to modulate intermediate features, encouraging the network to make more reliable decisions near uncertain crack boundaries. Beyond improving segmentation, the variance map also offers an interpretable diagnostic cue, indicating where uncertainty concentrates and supporting error analysis and model refinement.

\section{Methodology}

\begin{figure*}
\centering
\includegraphics[width=12 cm]{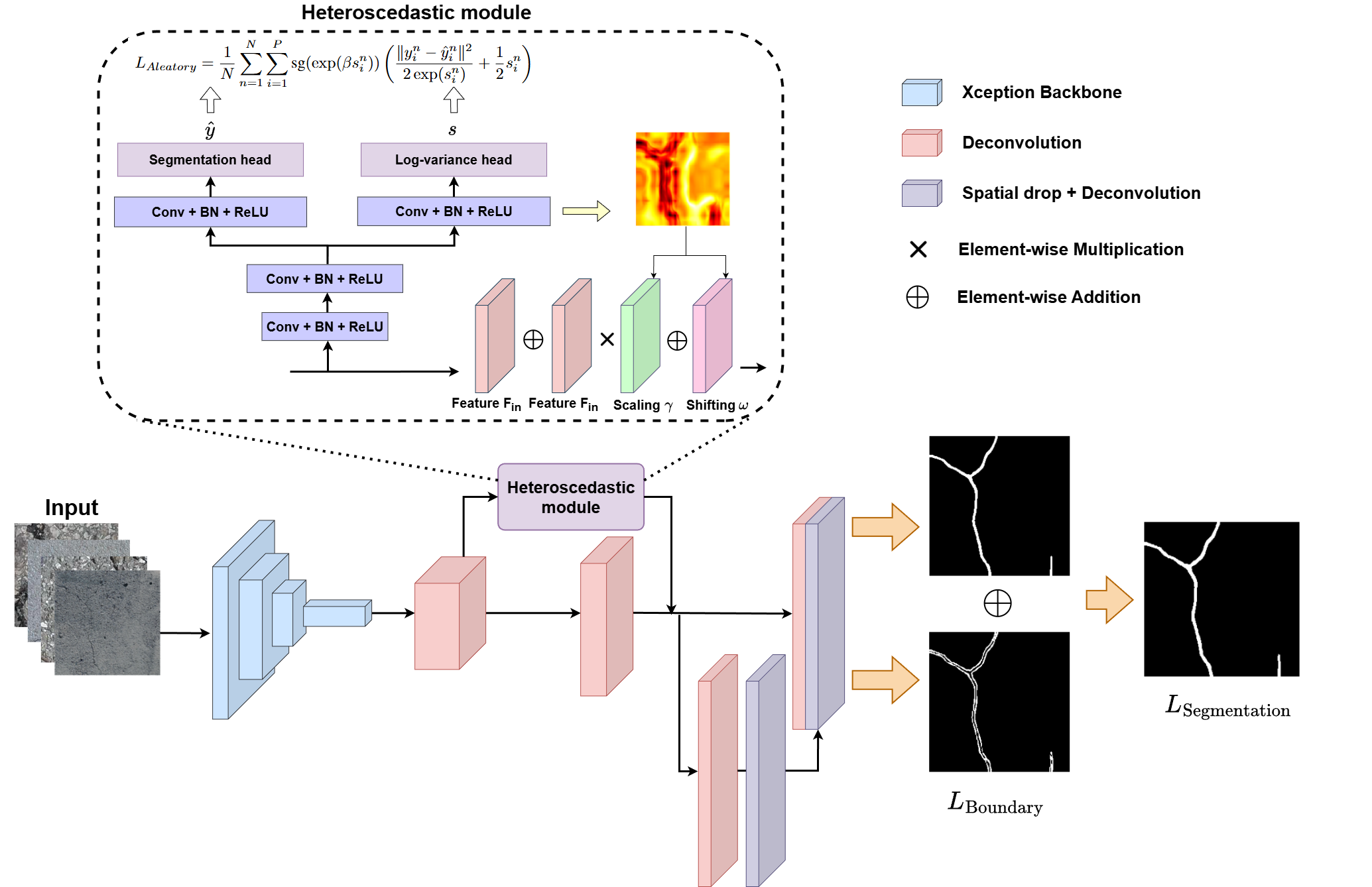}
\caption{Overview of the proposed UnGAP architecture. The framework employs an Xception-based encoder-decoder structure. The core innovation is the Heteroscedastic Module (top), which establishes a closed-loop mechanism between uncertainty estimation and feature learning. This module predicts pixel-wise log-variance ($s$) alongside the segmentation map ($\hat{y}$), optimized via the $\beta$-NLL Aleatoric loss ($L_{Aleatory}$) to prevent gradient attenuation in hard samples. Crucially, the estimated uncertainty serves as a visual prompt for the Uncertainty-Prompted Feature Modulator (UPFM). The UPFM generates dynamic scaling ($\gamma$) and shifting ($\omega$) parameters to spatially calibrate the input features ($F_{in}$) via affine transformation (element-wise multiplication and addition). Finally, the calibrated features are processed through a dual-branch detection head to output both boundary and segmentation predictions, ensuring precise delineation of crack edges.}
\label{overall framework}
\end{figure*}

The proposed UnGAP framework is designed as encoder-decoder architecture for real-time performance, which illustrated in Fig.~\ref{overall framework}. It comprises three components: a backbone encoder, a Heteroscedastic Module equipped with the Uncertainty-Prompted Feature Modulator (UPFM), and a simle but effective Boundary-aware Detection Head. 

The backbone network extract hierarchical feature representations as the encoder from the input image to capture both local textures and gradients. The subsequent decoder pathway focuses on restoring spatial resolution and refining segmentation details. Distinguishing itself from conventional designs, the Heteroscedastic Module does not only output uncertainty for manual analysis; instead, it establishes a closed-loop mechanism. It explicitly quantifies pixel-wise uncertainty using a $\beta$-NLL constrained objective and utilizes this uncertainty map as an active visual prompt. This prompt drives the UPFM to dynamically calibrate feature distributions, specifically targeting ambiguous regions where the model lacks confidence. Finally, the Boundary-aware Detection Head uses boundary-oriented learning to enforce geometric consistency along crack edges, effectively constraining predictions and reducing redundancy. The details of each module are elaborated in the following sections.

\subsection{Backbone module}

The model uses a pre-trained model based on CNN as the backbone network for feature extraction, serving as an essential component for learning high-level representations. The UnGAP can be regarded as a encoding-decoding structure, where the backbone module functions as the encoder. This encoder extracts hierarchical features from the input image, capturing both local and global context. The decoding part of the model takes these encoded features and refines them through upsampling layers and other techniques to generate detailed segmentation maps. In this module, we applied pre-trained Xception and ResNet-101 as our backbone.

\subsection{Heteroscedastic module}

The detailed framework of Heteroscedastic module is shown in Fig.~\ref{overall framework}. Aleatoric uncertainty allows the model to account for these variations by identifying features that may be ambiguous or noisy. By assigning different levels of uncertainty to different regions of the image, the model can distinguish between true cracks and similar-looking artifacts or noise. This means uncertainty varies spatially, as the probability map highlights regions where the data is noisier or more uncertain, showing that the uncertainty is not uniform across the image, we refer this phenomenon as heteroscedasticity.



To capture this characteristic, our Heteroscedastic module operates in parallel with the decoder, explicitly predicting a pixel-wise log-variance map $s$ that quantifies the model's lack of confidence at each pixel. Unlike conventional approaches that treat uncertainty map as simple output, our framework establishes a closed-loop mechanism that utilizes this map as a constructive signal. The predicted uncertainty serves a dual purpose in our architecture. First, it acts as an adaptive weight in the optimization process, balancing the learning signal via a modified loss function to prevent the model from overfitting to noise while ensuring that difficult samples are not neglected due to gradient attenuation. Second, and more importantly, it functions as an active visual prompt for the subsequent Feature Modulation, identifying ambiguous features that require dynamic calibration. The specific mechanisms for the loss function and the feature modulation are detailed in the following subsections.

\subsubsection{Aleatoric loss}

To enable the model to identify high-variance regions without falling into the optimization pitfalls of standard likelihood maximization, we adopt the $\beta$-NLL loss function~\cite{Seitzer2022}. Below, we explain the inference process and the formulation of this loss.

We assume that the observed label $y_i^n$ at every pixel is noisy. To capture the aleatoric uncertainty arising from this noise, we model the labels as drawing from a Gaussian distribution with the model's predicted mean $\hat{y}$ and variance $\sigma^2$:
\begin{equation}
y \sim \mathcal{N}(\hat{y}, \sigma^2)
\end{equation}
Standard uncertainty estimation typically employs Maximum Likelihood Estimation (MLE) to learn the parameters $\theta$. Minimizing the Negative Log-Likelihood (NLL) is equivalent to maximizing the likelihood:
\begin{equation}
\mathcal{L}_{\text{NLL}}(\theta) = -\log p(y | \hat{y}, \sigma) = \frac{1}{2\sigma^2} \| y - \hat{y} \|^2 + \frac{1}{2} \log \sigma^2 + \text{constant}
\label{eq:standard_nll}
\end{equation}
In this formulation, the gradient of the loss with respect to the prediction $\hat{y}$ is scaled by the inverse variance $1/\sigma^2$. As discussed in previous sections, this creates a structural pathology: the model can minimize the loss by simply predicting large variance $\sigma^2$ for difficult samples (e.g., blurred cracks), which attenuates the gradients and causes the model to effectively ``ignore'' these regions during training~\cite{Seitzer2022}.

To counteract this gradient attenuation while maintaining probabilistic consistency, we employ the $\beta$-NLL loss. This formulation introduces a stop-gradient weighted term to balance the learning signal. We define the log-variance $s = \log(\sigma^2)$ as the direct network output for numerical stability. The $\beta$-NLL loss is defined as:

\begin{equation}
\mathcal{L}_{\beta\text{-NLL}} = \text{sg}(\sigma^{2\beta}) \left( \frac{1}{2}\exp(-s) \| y - \hat{y} \|^2 + \frac{1}{2}s \right)
\end{equation}

Where $\text{sg}(\cdot)$ denotes the stop-gradient operator, meaning the term $\sigma^{2\beta}$ is treated as a constant during back-propagation and does not contribute to the gradient computation for $\sigma$. The hyperparameter $\beta \in [0, 1]$ controls the degree of gradient attenuation. When $\beta = 0$, it reverts to the standard NLL; when $\beta = 1$, the gradient magnitude mimics the standard MSE loss, preventing the model from under-fitting high-uncertainty regions. Following Seitzer et al., we set $\beta = 0.5$, which strikes a balance between robust variance estimation and effective mean fitting.

Eventually, the final Aleatoric loss function aggregated over the dataset is defined as:

\begin{equation}
L_{Aleatory} = \frac{1}{N} \sum_{n=1}^{N} \sum_{i=1}^{P} \text{sg}(\exp(\beta s_i^n)) \left( \frac{\| y_i^n - \hat{y}_i^n \|^2}{2\exp(s_i^n)} + \frac{1}{2}s_i^n \right)
\end{equation}

Where $N$ is the number of images, $P$ is the number of pixels, and $s_i^n$ is the predicted log-variance. By using this re-weighted loss, UnGAP ensures that regions with high aleatoric uncertainty (large $s$) are identified but not ignored. The uncertainty map $s$ is then utilized by our Heteroscedastic Module and UPFM to guide feature refinement, as visualized in Fig.~\ref{sample uncertainty}.

\subsubsection{Uncertainty-Prompted Feature Modulation}

Crack segmentation presents unique challenges due to the lack of distinct global geometric shapes, relying heavily on fine-grained local textures and gradients. Consequently, in regions exhibiting high aleatoric uncertainty—such as blurred boundaries or complex backgrounds—the extracted features often suffer from ambiguity, residing dangerously close to the classifier's decision boundary. To address this, we propose the Uncertainty-Prompted Feature Modulator (UPFM). This mechanism draws inspiration from Feature-wise Linear Modulation (FiLM)~\cite{Perez2018}, a framework originally designed to inject multi-modal context (e.g., text embeddings) into visual models. In our work, we repurpose this concept to allow the network to adaptively "reshift" and "rescale" the feature space based on the model's own confidence level.

Crucially, our design provides an architectural solution to the structural pathology in heteroscedastic regression identified by Seitzer et al. and Wong-Toi et al.~\cite{Seitzer2022, WongToi2023}. In standard uncertainty-aware models, the loss gradient is weighted by the inverse variance ($1/\sigma^2$). This creates a perverse incentive where the model learns to predict high uncertainty for hard samples simply to attenuate the gradient and lower the NLL loss, effectively ``ignoring" difficult regions (undersampling). However, UPFM mitigate this question in different perspective. Instead of treating high uncertainty as a passive weight that suppresses the learning signal, UPFM utilizes the heteroscedastic map as an active visual prompt. When the model predicts high uncertainty for a pixel, it does not dampen the signal; conversely, it triggers the UPFM to generate stronger affine transformation parameters. Since the segmentation loss back-propagates through this modulation layer, the model is forced to use the high-uncertainty signal to rectify the features and minimize the segmentation error. Thus, we convert the ``high variance" from a signal of suppression into a prompt for calibration, ensuring that hard samples receive adaptive processing rather than being neglected.

Specifically, the UPFM takes the heteroscedastic map $h$ as input and generates pixel-wise affine transformation parameters: a scaling factor $\gamma$ and a shifting factor $\omega$. This is achieved through a lightweight convolutional generator consisting of two $1\times1$ convolutional layers with ReLU activation, ensuring real-time efficiency. The process can be formulated as:

\begin{equation}
    [\gamma, \omega] = \text{Conv}_{1\times1}\left(\delta\left(\text{Conv}_{1\times1}(h)\right)\right)
\end{equation}

Where $\delta$ denotes the ReLU activation function. The generated parameters $\gamma, \omega \in \mathbb{R}^{C \times H \times W}$ have the same spatial resolution as the input feature map $F_{in}$. The feature modulation is then applied via a conditional affine transformation:

\begin{equation}
    F_{refined} = F_{in} \odot (1 + \gamma) + \omega
\end{equation}

Where $\odot$ represents element-wise multiplication. This modulation mechanism offers a geometric interpretation in the feature space:
\begin{itemize}
    \item \textbf{Dynamic Scaling ($\gamma$):} In regions with high uncertainty, the network learns to predict a larger $\gamma$ to amplify feature variance. This acts as a learnable "contrast enhancement" or "sharpening" filter, making crack textures more distinct from the background.
    \item \textbf{Dynamic Shifting ($\omega$):} The shift parameter $\omega$ allows the model to translate the feature distribution. This effectively pushes ambiguous features away from the decision boundary towards the correct class cluster (crack or background), correcting the feature representation rather than just filtering it.
\end{itemize}

Finally, UPFM uses uncertainty as a prompt to switch its processing logic, applying feature rectification in ambiguous crack boundaries while preserving smooth representations in confident backgrounds.

\subsection{Boundary-aware Detection head}

Although pixel-wise uncertainty provides strong support for the identification of cracks, the precise delineation of cracks largely relies on clear boundary lines. Boundary-aware detection head operates on a simple dual-branch network; 1) Boundary Net is tasked to learn and predict boundaries explicitly and guided by boundary loss function $L_{\text{Boundary}}$. 2) Segmentation Net generate the segmentation map without loss function. Eventually, the outputs of two branches are added in element-wise, and the result is guided by segmentation loss $L_{\text{Segmentation}}$. 

If the segmentation result is satisfactory, then the boundary information should not affect it. However, in incorrectly segmented regions, the boundary information will play a guiding role and improve the discriminative ability of the model in the boundary region by enhancing the gradient signal at the boundary. This enhancement help the model to correct the misclassified pixels more accurately, avoid the blurring of the crack boundary, and improve the accuracy and stability of the overall segmentation

We use boundary-aware segmentation to capture the morphology of cracks due to several reasons. Firstly, crack morphology is highly irregular and lacks a fixed global structure. Therefore, the boundary information of the crack determines the semantic integrity of the crack more than the overall shape and trajectory. This aligns with our focus on local characteristics, as the boundary is the ultimate local determinant of a crack's presence. Secondly, Building cracks may appear on the surfaces of various concrete materials or even non-concrete materials, which have large differences in the background texture. Therefore, boundary detection mainly focuses on gradient information, which is the pixel change between the crack and its background, rather than the absolute color or texture of the crack region. By focusing on gradients, sensitivity to different backgrounds can be reduced, thereby maintaining the model's stable detection performance. Finally, boundary detection method is simple and effective and easy to implement, which is more suitable for real-time tasks.

\subsection{Overall Loss function}

The overall loss function can be defined as below:

\begin{equation}
L_{\text{Final}} = w_1 \cdot L_{\text{Aleatory}} + w_2 \cdot L_{\text{Boundary}} + w_3 \cdot L_{\text{Segmentation}}
\end{equation}

Where the $L_{\text{Aleatory}}$ denote the Aleatoric loss. $L_{\text{Segmentation}}$ and $L_{\text{Boundary}}$ denote as segmentation and boundary loss respectively. These two losses functions are formulated as typical dice loss function. This is due to Dice loss is more effective for small classes~\cite{Sudre2017}, and in cracks and boundary of crack, it only occupy a small portion of the image. Each term in the loss function is weighted by a corresponding coefficient to ensure an appropriate balance between addressing uncertainty, object segmentation and boundary detection.

\section{Experiment}
\subsection{Dataset}

The work is experimented using various datasets, including Crack500, CrackSeg9k, CrackVision12K, CrackTree200 and Crack Forest Dataset (CFD). Table~\ref{datasets_summary} shows the details of these datasets.

\begin{table}[hbt!]
\caption{Summary of datasets used in the experiments.}
\centering
\resizebox{10cm}{!}{%
\begin{tabular}{|l|c|c|c|c|}
\hline
\textbf{Dataset} & \textbf{Resolution}  & \textbf{Training} & \textbf{Validation} & \textbf{Testing}\\
\hline
\textbf{Crack500~\cite{Yang2020}} & 360 x 640 & 1,896 & 348 & 1,124 \\
\hline
\textbf{CrackTree200~\cite{Qin2012}} & 800 x 600 & 164 & 21 & 21 \\
\hline
\textbf{CFD~\cite{Shi2016}} & 480 x 320 & 96 & 11 & 11 \\
\hline
\textbf{CrackSeg9k~\cite{Kulkarni2023}} & 448 x 448 & 9,603 & 803 & 802 \\
\hline
\textbf{CrackVision12K~\cite{goo2024hybridsegmentor}} & 256 x 256 & 9600 & 1200 & 1200\\
\hline
\end{tabular}
}
\label{datasets_summary}
\end{table}

\begin{table*}[htbp]
\caption{Performance comparison with different baselines on five crack datasets. * represent results retrieved from~\cite{chen2024}, others are our experiment results}
\centering
\resizebox{\columnwidth}{!}{
\begin{tabular}{lccccccccccccccc}  
\toprule 
\multirow{2}{*}{Method} & \multicolumn{3}{c}{CrackTree200} & \multicolumn{3}{c}{Crack500} & \multicolumn{3}{c}{CFD} & \multicolumn{3}{c}{CrackVision12K} & \multicolumn{3}{c}{CrackSeg9K} \\
\cmidrule(lr){2-4} \cmidrule(lr){5-7} \cmidrule(lr){8-10} \cmidrule(lr){11-13} \cmidrule(lr){14-16}
& Precision & Recall & F1 & Precision & Recall & F1 & Precision & Recall & F1 & Precision & Recall & F1 & Precision & Recall & F1 \\
\midrule 
UNet~\cite{Ronneberger2015} &  79.16$^*$ & 78.95$^*$ & 78.42$^*$ & 62.22$^*$ & 68.85$^*$ & 61.83$^*$ & \textbf{69.73} & 63.60 & 66.52 & 77.55 & 79.32 & 78.13 &  74.29 &  70.93 &  72.81 \\
Crackformer~\cite{Huajun2021} & 84.13$^*$ & 81.93$^*$ & 83.42$^*$ & 69.13$^*$ & 66.24$^*$ & 64.75$^*$ & 66.77 & 71.76 & 69.18 & 82.85 & 83.98 & 83.28 & 75.72  &  73.97 &  74.90 \\
JTFN~\cite{Cheng2021} & 85.87$^*$ & 82.58$^*$ & 84.19$^*$ & 68.81$^*$ & 69.06$^*$ & 65.76$^*$ & 68.79 & 69.17 & 69.04 & 84.05 & 85.77 & 85.53 & 76.85  &  71.18 &  73.26 \\
JTFN + CIRL~\cite{chen2024} & \textbf{87.62$^*$} & 83.92$^*$ & 86.53$^*$ & 70.32$^*$ & 69.93$^*$ & 67.62$^*$ & - & - & - & - & - & - & - & - & - \\
probabilistic U-Net~\cite{Kohl2018} & 75.53 & 77.92 & 76.41 & 64.95 & 67.64 & 66.52 & 42.61 & 47.03 & 44.70 & 64.17 & 69.33 & 66.50 & 56.17 & 64.67 & 59.98  \\

PCSD~\cite{Mengxue2026}
& 76.84 & 79.36 & 78.22 
& 62.10 & 66.35 & 64.88 
& 63.71 & 65.02 & 64.25 
& 75.84 & 81.62 & 78.86 
& 68.12 & 74.20 & 70.35 \\

B-BACN~\cite{Rahul2023} & 84.73 & 81.99 & 82.21 & 65.44 & 63.29 & 64.06 & 68.56 & 64.61 & 67.63 & 78.47 & 76.15 & 77.49 & 67.12 & 66.09 & 66.53\\
TernausNet~\cite{Ketson2025} & 83.72 & 84.70 & 84.14 & 65.02 & 67.71 & 66.87 & 57.13 & 54.25 & 58.96 & 73.50 & 76.39 & 75.91 & 69.21 & 66.23 &  66.57 \\

SelectSeg~\cite{Chen2025}
& 84.46 & 82.15 & 83.31 
& 68.76 & 66.82 & 65.10 
& 67.10 & 71.21 & 69.05 
& 82.43 & 84.31 & 83.22 
& 75.28 & 74.36 & 74.77 \\

CCDM~\cite{Zbinden2023} & 77.35 & 78.94 & 78.51 & 62.39 & 66.69 & 65.16 & 64.26 & 65.40 & 64.69 & 76.39 & 82.47 & 79.34 & 68.87 & \textbf{75.25} &  70.99\\

\midrule 
\textbf{UnGAP(Xception)} & 85.93 & \textbf{88.61} & \textbf{87.04} & 70.12 & \textbf{72.05} & \textbf{71.33} & 69.42 & \textbf{72.04} & \textbf{70.15 } & \textbf{85.32} & \textbf{88.01} & \textbf{87.49} & \textbf{ 77.94 } &  72.60  & \textbf{ 75.19 } \\
\bottomrule  
\end{tabular}
}
\label{baseline performances}
\end{table*}


\begin{figure*}
	\centering
\includegraphics[width=12 cm]{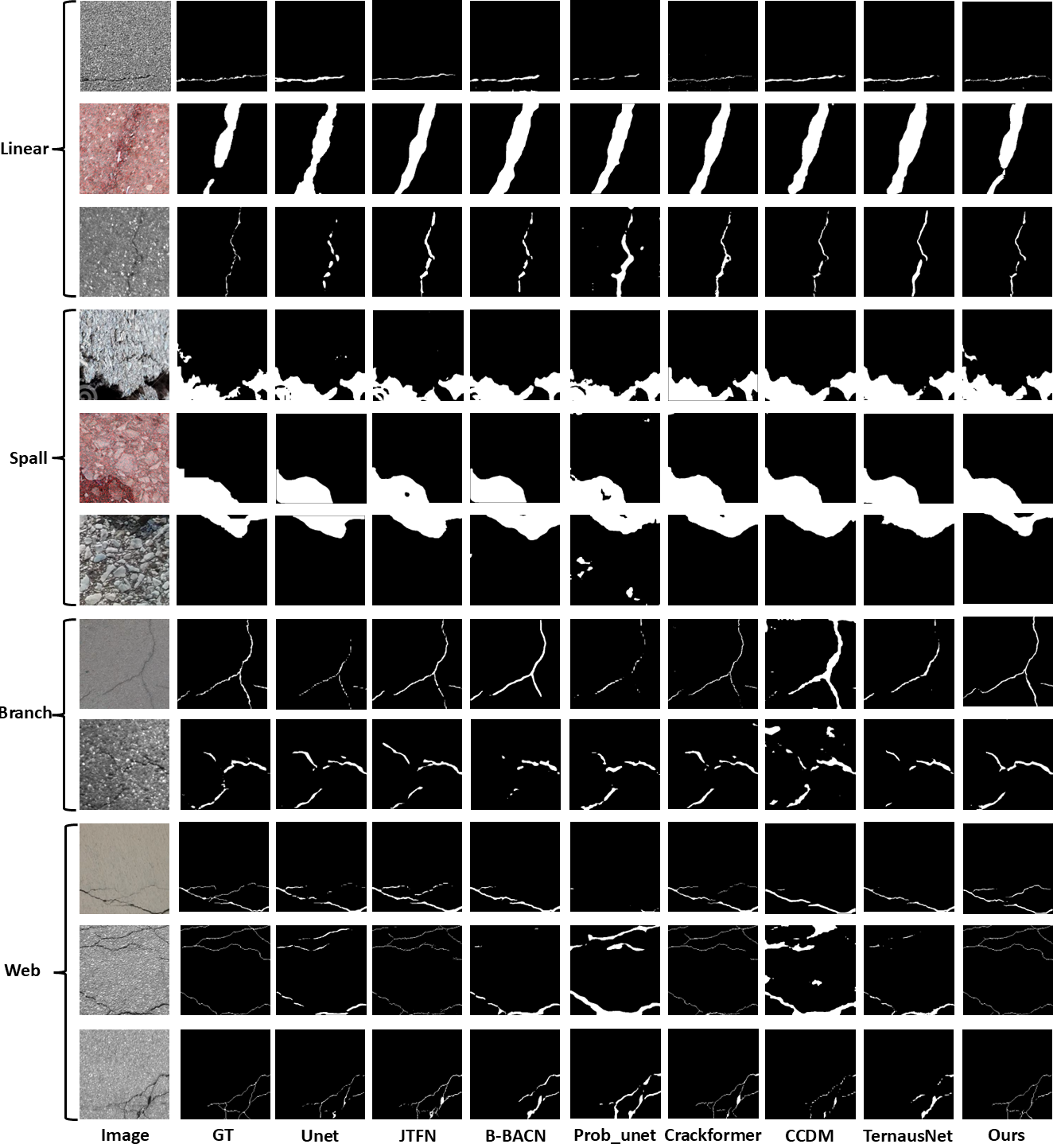}
\caption{ Visual comparison of segmentation results on the crackvision12k dataset. The rows display different crack topologies (Linear, Spall, Branch, and Web), comparing our method against seven sota baselines. }
\label{sample result}
\end{figure*}


\subsection{Implementation detail}

\textbf{Model setting}. In UnGAP, we use Xception and resnet-101 as backbone network. In deconvolution layers, for light-weight design, we set bilinear interpolation and 1×1 convolution to obtain the deconvolution outcome. Every deconvolution layer is constituted by 1×1 convolution, max pooling, Relu activation and one more 1×1 convolution. 

\textbf{Training}. In all experiments, to enrich the data volume, we perform random crop to set the size of input image as 400 × 400, and select one of the argumentation methods randomly. The random argumentation methods include: Random Rotation, Random Brightness Contrast and Gaussian Blur. In hyper-parameter setting, we set the size of input image as 400 × 400, learning rate as 0.0002, and loss weight $w1$=0.87, $w2$=0.13, $w3$=0.001. Because the aleatoric loss is based on negative log-likelihood with high loss value, and the range of dice loss is from 0 to 1. So if we do not set a small value to $w3$, the model will pay more attention on variance identification instead of final segmentation. Lastly we optimized the model training with Adam optimizer for 1500 epoch, and with batch size of 32. All experiments are conducted using NVIDIA A100 Tensor Core GPU.

\textbf{Evaluation metrics}. To evaluate the pixel-wise performance of the proposed method and other baseline methods in different dataset, following existing work~\cite{Seyedhosseini2013, Hu2019, chen2024}, we set micro F1 score, precision and recall as our evaluation metrics. F1 score is the harmonic average of recall and precision, offering a better reflection of balanced performance. It can be expressed as $\text{F1 Score} = 2 \times \frac{\text{Precision} \times \text{Recall}}{\text{Precision} + \text{Recall}}$. These metrics provide more detailed and meaningful evaluation to recognize data imbalance and focus on label class.

\subsection{Result analysis}


In this section, we provide a comprehensive evaluation of the proposed UnGAP by comparing it against state-of-the-art baseline methods. All test images were standardized to a scale of $400 \times 400$ without the application of multi-scale testing or multi-crop evaluation to ensure a fair comparison of raw model capability. We categorized the baselines into two groups: conventional deterministic methods (U-Net, Crackformer, JTFN, JTFN+CIRL) and uncertainty-aware stochastic methods (Probabilistic U-Net, B-BACN, TernausNet, CCDM).

As presented in table~\ref{baseline performances}, UnGAP demonstrates consistent superiority across all five datasets, achieving the highest F1-scores in nearly every benchmark. This indicates our method's strong generalization ability across varying resolutions and texture complexities. Specifically, on the CrackVision12K dataset, which contains complex pavement backgrounds, UnGAP achieves a remarkable performance leap. Compared to JTFN, the previous state-of-the-art performer, our method improves the F1-score by 5.09\%. Furthermore, when compared to stochastic methods like Probabilistic U-Net and B-BACN, UnGAP exhibits a significant advantage. For instance, on the CrackSeg9k dataset, UnGAP outperforms B-BACN by over 8\% in F1-score. UnGAP maintains high values in both metrics (90.30\% Precision and 90.99\% Recall on CrackVision12K), proving that our UPFM effectively helps the model recover ``hard" samples that are typically ignored by other models.

\begin{figure*}
	\centering
\includegraphics[width=12 cm]{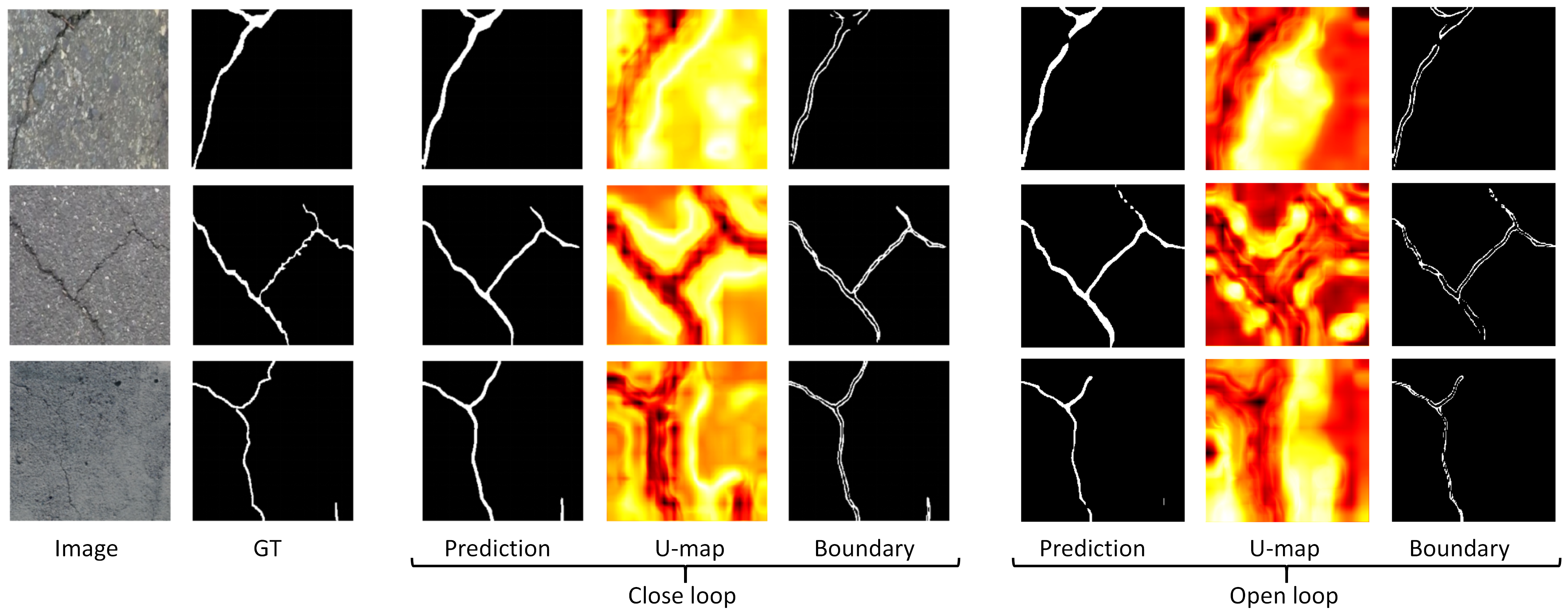}
\caption{Comparison of uncertainty behaviors between the Closed-loop (Ours) and Open-loop settings. The Closed-loop mechanism generates focused uncertainty maps that tightly align with crack boundaries, whereas the Open-loop baseline produces spatially diffuse uncertainty distributions.}
\label{sample uncertainty}
\end{figure*}   

To intuitively demonstrate the effectiveness of UnGAP, we present qualitative comparisons in Fig.~\ref{sample result}. The visual results highlight the robustness of our model in handling diverse crack typologies compared to baseline methods. A notable trend observed in Fig.~\ref{sample result} is the difference in prediction clarity. Stochastic segmentation methods (e.g., Prob U-Net) often yield predictions with blurred boundaries or excessive noise, effectively trading spatial precision for uncertainty estimation. In contrast, UnGAP generates sharper, well-defined segmentation maps that closely align with the Ground Truth. By actively leveraging the heteroscedastic uncertainty to modulate features rather than merely outputting it, our model successfully suppresses background artifacts while maintaining the structural continuity of the cracks, proving its effectiveness even in challenging visual scenarios.

Real-time performance is a critical requirement for structural health monitoring. Table~\ref{speed performance} details the trade-off between model complexity and inference speed. While Crackformer achieves competitive accuracy, its heavy reliance on Transformer blocks results in high computational cost and a low inference speed of 19.26 FPS, which is insufficient for real-time applications. Similarly, JTFN suffers from a massive parameter size (237.79 MB). In contrast, UnGAP achieves an exceptional balance. Benefiting from the lightweight Xception backbone and efficient design of the UPFM, our model operates at 147.79 FPS, which is approximately 7.6 times faster than Crackformer and nearly 2 times faster than the widely used U-Net. Moreover, with only 69.38 GFLOPs, UnGAP requires significantly less computational power than U-Net and B-BACN, making it highly suitable for deployment on edge devices with limited resources.

\begin{table}[htbp]
\caption{Parameter size, inference speed and FPS comparison with different baseline methods.}
\centering{
\resizebox{10cm}{!}{
\begin{tabular}{lrrrr}  
\toprule
Model & Param(MB) & GFLOPs & FPS \\
\midrule
U-Net~\cite{Ronneberger2015} & 141.00 & 355.93 & 78.85 \\
prob U-Net~\cite{Kohl2018} & 19.09 & 68.25 & 77.04 \\
Crackformer~\cite{Huajun2021} & 18.90 & 107.78 & 19.26 \\
JTFN~\cite{Cheng2021} & 237.79 & 78.63 & 41.79 \\
SelectSeg~\cite{Chen2025} & 23.18 & 82.90 & 14.55 \\
B-BACN~\cite{Rahul2023} & 181.56 & 101.04 & 88.43 \\
TernausNet~\cite{Ketson2025} & 172.48 & 261.95 & 73.32 \\
UnGAP(xception) & 96.01 & 69.38 & 147.79 \\
\bottomrule
\end{tabular}
}
\label{speed performance}
}
\end{table}

\subsection{Ablation study}

\subsubsection{Module test}

In this section, we conduct a comprehensive ablation study on the CrackSeg-9k dataset to investigate the contribution of each individual component in the proposed UnGAP framework. Specifically, we evaluate the impact of the Heteroscedastic Module (HM), the Uncertainty-Prompted Feature Modulator (UPFM), and the Boundary-aware Detection Head (BDH). The quantitative results are summarized in table~\ref{ablation_study}.

\begin{table}[t]
\centering
\caption{Ablation study on the validation set of CrackSeg-9k. HM: Heteroscedastic Module; UPFM: Uncertainty-Prompted Feature Modulator; BDH: Boundary-aware Detection Head. The Baseline denotes the standard encoder-decoder network.}
\label{ablation_study}
\setlength{\tabcolsep}{7pt}
\resizebox{12cm}{!}{
\begin{tabular}{c|ccc|ccc}
\hline
\multirow{2}{*}{Exp.} & \multicolumn{3}{c|}{Components} & \multicolumn{3}{c}{Metrics (\%)} \\ \cline{2-7} 
 & HM & UPFM & BDH & Precision & Recall & F1-Score \\ \hline
1 & - & - & - & 68.11 & 54.17 & 61.38 \\
2 & \checkmark & - & - & 70.25 & 65.30 & 67.68 \\ 
3 & \checkmark & \checkmark & - & 73.40 & 74.66 & 74.02 \\ 
4 & \checkmark & \checkmark & \checkmark & \textbf{77.94} & \textbf{72.60} & \textbf{75.19} \\ \hline
\end{tabular}
}
\end{table}


\begin{enumerate}
    \item Effectiveness of the Baseline: The first row of Table IV represents our baseline model, which utilizes a standard Xception encoder-decoder architecture optimized with conventional segmentation loss. It achieves an F1-score of 61.38\%, serving as the reference point for evaluating our proposed modules.

    \item Impact of Uncertainty Estimation (HM): In the second experiment, we integrate the Heteroscedastic Module to predict pixel-wise aleatoric uncertainty, trained with the $\beta$-NLL loss. While this allows the model to capture variance maps, simply treating uncertainty as an auxiliary output yields only moderate improvements. This aligns with our hypothesis that merely identifying uncertainty without feeding it back into the feature learning process is insufficient to fully resolve segmentation ambiguities.

    \item Importance of Closed-loop Modulation (UPFM): In this experiment, we incorporate the UPFM to boost the F1-score to 74.02\%. This result fundamentally validates the superiority of our closed-loop design over conventional uncertainty-aware methods. The significant performance improvement observed in the third experiment validates the effectiveness of the closed-loop design. In the standard approach (Exp. 2), the uncertainty map serves solely as an output used for loss weighting or post-processing analysis. Consequently, the uncertainty estimation is isolated from the feature learning process. Although the model identifies regions with high variance, it lacks a mechanism to utilize this information to improve feature representation during the forward pass. This often results in the model reducing the loss weight in difficult regions rather than refining the features.

    \item Refinement with Boundary-aware Head (BDH): Finally, adding the Boundary-aware Detection Head brings the performance to its peak, achieving an F1-score of 75.19\%. While the improvement in F1-score is incremental compared to the UPFM, the BDH plays a crucial role in geometric precision. It explicitly constrains the model to pay attention to gradient changes, resulting in cleaner and more compact segmentation boundaries.

\end{enumerate}

The ablation analysis confirms that while uncertainty estimation and boundary detection are beneficial, the UPFM and our end-loop design for uncertainty map is the dominant factor driving the performance of UnGAP. It effectively bridges the gap between uncertainty analysis and feature representation learning.

\subsubsection{Uncertainty Map Behavior}

Beyond the quantitative performance, we further investigate how the closed-loop mechanism affects the morphological characteristics of the predicted uncertainty maps. We compare the uncertainty outputs generated by the model with only the Heteroscedastic Module and boundary head against the full UnGAP framework (Exp. 4) in Fig.~\ref{sample uncertainty}.

In the absence of the feedback loop, the predicted uncertainty maps tend to be spatially diffuse and coarse. Since the uncertainty is utilized primarily for loss attenuation (via the $\frac{1}{2\sigma^2}$ term in the $\beta$-NLL loss), the model learns to predict high variance over broad regions surrounding the cracks to minimize the penalty from hard samples. Consequently, the uncertainty map acts as a general "mask" for ambiguous areas rather than a precise indicator of pixel-level difficulty. In contrast, when UPFM is integrated, the uncertainty maps become significantly more fine-grained and tightly aligned with the crack boundaries. This shift occurs because the uncertainty now serves as a functional trigger for feature modulation. The model learns to pinpoint high uncertainty specifically at the decision boundaries where feature shifting and scaling are most critical for distinguishing the crack from the background.

This morphological transition demonstrates that the closed-loop design changes the role of uncertainty estimation. Instead of only reflecting data ambiguity to dampen the loss, the uncertainty map in UnGAP evolves into a localized attention mechanism, identifying and rectifying specific features to improve segmentation confidence.

\section{Conclusion}

In this work, we presented UnGAP, a real-time framework for crack segmentation designed for scenarios with strong aleatoric uncertainty, where crack regions often exhibit weak contrast, irregular morphology, fragmented continuity, and ambiguous boundaries against complex backgrounds. Unlike semantic objects with relatively stable global structure, cracks are mainly characterized by subtle local textures and boundary gradients, which makes reliable dense prediction particularly challenging. To address this issue, we introduced three tightly coupled components: a Heteroscedastic Module (HM) trained with a $\beta$-NLL objective to estimate calibrated pixel-wise uncertainty, an Uncertainty-Prompted Feature Modulator (UPFM) that feeds the estimated uncertainty back into decoder features for adaptive refinement, and a lightweight Boundary-aware Detection Head (BDH) that further improves localization around thin and discontinuous crack boundaries. Extensive experiments on multiple crack benchmarks demonstrate that UnGAP consistently improves segmentation accuracy over strong real-time and lightweight baselines while maintaining favorable computational efficiency. The ablation results further show that HM, UPFM, and BDH provide complementary gains, confirming the effectiveness of uncertainty estimation, uncertainty-guided modulation, and boundary-aware supervision within a unified framework. Beyond the empirical improvements, this work also provides a broader insight for dense prediction under ambiguity: uncertainty should not be viewed only as a passive confidence score for post-hoc analysis, but can also be exploited as an active signal to guide feature refinement during inference. This perspective is valuable for practical crack inspection, where uncertain regions often coincide with structurally important but visually ambiguous defects. By explicitly closing the loop between uncertainty estimation and feature modulation, UnGAP offers a simple yet effective way to improve robustness without sacrificing real-time deployment capability.

Nevertheless, several limitations remain. First, the effectiveness of the closed-loop refinement mechanism still depends on the quality of the estimated uncertainty maps; if uncertainty is poorly calibrated, the benefit of UPFM may be reduced. Second, severe domain shifts, such as changes in material type, illumination, surface contamination, or acquisition devices, may introduce biased uncertainty patterns and degrade generalization. Third, very high-resolution imagery and long-range crack structures can still increase memory usage and latency in practical deployment. Future work will therefore focus on improving uncertainty calibration under domain shift, developing stronger yet more stable uncertainty-guided modulation strategies, and extending the framework toward richer uncertainty modeling and large-scale structural inspection scenarios.


\section*{CRediT}

\textbf{Conghui Li}: Conceptualization, Methodology, Software, Formal analysis, Investigation, Data curation, Validation, Visualization, Writing -- original draft. \textbf{Huanyu He}: Validation, Formal analysis, Writing -- review \& editing. \textbf{Xin Wang}: Supervision, Writing -- review \& editing. \textbf{Weiyao Lin}: Methodology, Supervision, Writing -- review \& editing. \textbf{Chern Hong Lim}: Conceptualization, Methodology, Supervision, Project administration, Resources, Funding acquisition, Writing -- review \& editing.

\section*{Declaration of competing interest}
The authors declare that they have no known competing financial interests or personal relationships that could have appeared to influence the work reported in this paper.

\section*{Data availability}

The data used in this study are publicly available.


\bibliographystyle{IEEEtran}

\bibliography{cas-refs}

@inproceedings{Dosovitskiy2020,
title={An Image is Worth 16x16 Words: Transformers for Image Recognition at Scale},
author={Alexey Dosovitskiy and Lucas Beyer and Alexander Kolesnikov and Dirk Weissenborn and Xiaohua Zhai and Thomas Unterthiner and Mostafa Dehghani and Matthias Minderer and Georg Heigold and Sylvain Gelly and Jakob Uszkoreit and Neil Houlsby},
booktitle={International Conference on Learning Representations},
year={2021},
url={https://openreview.net/forum?id=YicbFdNTTy}
}

@INPROCEEDINGS{Huajun2021,
  author={Liu, Huajun and Miao, Xiangyu and Mertz, Christoph and Xu, Chengzhong and Kong, Hui},
  booktitle={2021 IEEE/CVF International Conference on Computer Vision (ICCV)}, 
  title={CrackFormer: Transformer Network for Fine-Grained Crack Detection}, 
  year={2021},
  volume={},
  number={},
  pages={3763-3772},
  doi={10.1109/ICCV48922.2021.00376}}

@InProceedings{Zhao2018,
author = {Zhao, Hengshuang and Zhang, Yi and Liu, Shu and Shi, Jianping and Loy, Chen Change and Lin, Dahua and Jia, Jiaya},
title = {PSANet: Point-wise Spatial Attention Network for Scene Parsing},
booktitle = {Proceedings of the European Conference on Computer Vision (ECCV)},
month = {September},
year = {2018}
}

@InProceedings{Ronneberger2015,
author="Ronneberger, Olaf
and Fischer, Philipp
and Brox, Thomas",
editor="Navab, Nassir
and Hornegger, Joachim
and Wells, William M.
and Frangi, Alejandro F.",
title="U-Net: Convolutional Networks for Biomedical Image Segmentation",
booktitle="Medical Image Computing and Computer-Assisted Intervention -- MICCAI 2015",
year="2015",
publisher="Springer International Publishing",
address="Cham",
pages="234--241",
isbn="978-3-319-24574-4"
}

@ARTICLE{Wang2021,
  author={Wang, Jingdong and Sun, Ke and Cheng, Tianheng and Jiang, Borui and Deng, Chaorui and Zhao, Yang and Liu, Dong and Mu, Yadong and Tan, Mingkui and Wang, Xinggang and Liu, Wenyu and Xiao, Bin},
  journal={IEEE Transactions on Pattern Analysis and Machine Intelligence}, 
  title={Deep High-Resolution Representation Learning for Visual Recognition}, 
  year={2021},
  volume={43},
  number={10},
  pages={3349-3364},
  doi={10.1109/TPAMI.2020.2983686}
}

@article{Yongshang2023,
title = {Real-time high-resolution neural network with semantic guidance for crack segmentation},
journal = {Automation in Construction},
volume = {156},
pages = {105112},
year = {2023},
issn = {0926-5805},
doi = {https://doi.org/10.1016/j.autcon.2023.105112},
author = {Yongshang Li and Ronggui Ma and Han Liu and Gaoli Cheng}
}

@article{Liang2017,
  author       = {Liang{-}Chieh Chen and
                  George Papandreou and
                  Florian Schroff and
                  Hartwig Adam},
  title        = {Rethinking Atrous Convolution for Semantic Image Segmentation},
  journal      = {CoRR},
  volume       = {abs/1706.05587},
  year         = {2017},
  url          = {http://arxiv.org/abs/1706.05587}
}

@article{Yahui2019,
title = {DeepCrack: A deep hierarchical feature learning architecture for crack segmentation},
journal = {Neurocomputing},
volume = {338},
pages = {139-153},
year = {2019},
issn = {0925-2312},
doi = {https://doi.org/10.1016/j.neucom.2019.01.036},
author = {Yahui Liu and Jian Yao and Xiaohu Lu and Renping Xie and Li Li},
}

@inproceedings{Baumgartner2019,
author = {Baumgartner, Christian F. and Tezcan, Kerem C. and Chaitanya, Krishna and H\"{o}tker, Andreas M. and Muehlematter, Urs J. and Schawkat, Khoschy and Becker, Anton S. and Donati, Olivio and Konukoglu, Ender},
title = {PHiSeg: Capturing Uncertainty in Medical Image Segmentation},
year = {2019},
isbn = {978-3-030-32244-1},
publisher = {Springer-Verlag},
address = {Berlin, Heidelberg},
url = {https://doi.org/10.1007/978-3-030-32245-8_14},
doi = {10.1007/978-3-030-32245-8_14},
booktitle = {Medical Image Computing and Computer Assisted Intervention – MICCAI 2019: 22nd International Conference, Shenzhen, China, October 13–17, 2019, Proceedings, Part II},
pages = {119–127},
numpages = {9},
location = {Shenzhen, China}
}

@article{Alex2015,
  author       = {Alex Kendall and
                  Vijay Badrinarayanan and
                  Roberto Cipolla},
  title        = {Bayesian SegNet: Model Uncertainty in Deep Convolutional Encoder-Decoder
                  Architectures for Scene Understanding},
  journal      = {CoRR},
  volume       = {abs/1511.02680},
  year         = {2015}
}

@inproceedings{Kohl2018,
 author = {Kohl, Simon and Romera-Paredes, Bernardino and Meyer, Clemens and De Fauw, Jeffrey and Ledsam, Joseph R. and Maier-Hein, Klaus and Eslami, S. M. Ali and Jimenez Rezende, Danilo and Ronneberger, Olaf},
 booktitle = {Advances in Neural Information Processing Systems},
 editor = {S. Bengio and H. Wallach and H. Larochelle and K. Grauman and N. Cesa-Bianchi and R. Garnett},
 pages = {},
 publisher = {Curran Associates, Inc.},
 title = {A Probabilistic U-Net for Segmentation of Ambiguous Images},
 url = {https://proceedings.neurips.cc/paper_files/paper/2018/file/473447ac58e1cd7e96172575f48dca3b-Paper.pdf},
 volume = {31},
 year = {2018}
}

@article{Tingbin2024,
title = {Experimental research on the degradation law of the bond performance between steel bars and concrete with rust expansion cracking},
journal = {Construction and Building Materials},
volume = {450},
pages = {138544},
year = {2024},
issn = {0950-0618},
doi = {https://doi.org/10.1016/j.conbuildmat.2024.138544},
author = {Tingbin Liu and Zhihan Xu and Tao Huang and Jiwu Yang and Yanbao Huang and Ning Xu and Congchao Pan and Lin Hua and Jiaxing Li}
}

@article{Shengyou2024,
title = {Research on the strength influence and crack evolution law of layered backfill based on macro and meso mechanical response},
journal = {Construction and Building Materials},
volume = {449},
pages = {138493},
year = {2024},
issn = {0950-0618},
doi = {https://doi.org/10.1016/j.conbuildmat.2024.138493},
author = {Shengyou Zhang and Wei Sun and Zhengmeng Hou and Aixiang Wu and Zhaoyu Li and Yingyu He and Bing Liu and Minggui Jiang and Shaoyong Wang}
}

@article{Khan2024,
title = {The influence of high temperature exposure on the tensile and cracking behavior of crimped-textile reinforced mortar composites (TRMs)},
journal = {Construction and Building Materials},
volume = {439},
pages = {137350},
year = {2024},
issn = {0950-0618},
doi = {https://doi.org/10.1016/j.conbuildmat.2024.137350},
author = {Khan Junaid and Nonna Algourdin and Zyed Mesticou and Gaochuang Cai and Amir {Si Larbi}}
}

@article{Armen2009,
title = {Aleatory or epistemic? Does it matter?},
journal = {Structural Safety},
volume = {31},
number = {2},
pages = {105-112},
year = {2009},
note = {Risk Acceptance and Risk Communication},
issn = {0167-4730},
doi = {https://doi.org/10.1016/j.strusafe.2008.06.020},
author = {Armen Der Kiureghian and Ove Ditlevsen},
}

@misc{Andrew2019,
      title={'In-Between' Uncertainty in Bayesian Neural Networks}, 
      author={Andrew Y. K. Foong and Yingzhen Li and José Miguel Hernández-Lobato and Richard E. Turner},
      year={2019},
      eprint={1906.11537},
      archivePrefix={arXiv},
      primaryClass={stat.ML},
      doi={https://arxiv.org/abs/1906.11537}, 
}

@ARTICLE{Yang2020,
  author={Yang, Fan and Zhang, Lei and Yu, Sijia and Prokhorov, Danil and Mei, Xue and Ling, Haibin},
  journal={IEEE Transactions on Intelligent Transportation Systems}, 
  title={Feature Pyramid and Hierarchical Boosting Network for Pavement Crack Detection}, 
  year={2020},
  volume={21},
  number={4},
  pages={1525-1535},
  doi={10.1109/TITS.2019.2910595}}

@InProceedings{Kulkarni2023,
author="Kulkarni, Shreyas
and Singh, Shreyas
and Balakrishnan, Dhananjay
and Sharma, Siddharth
and Devunuri, Saipraneeth
and Korlapati, Sai Chowdeswara Rao",
title="CrackSeg9k: A Collection and Benchmark for Crack Segmentation Datasets and Frameworks",
booktitle="Computer Vision -- ECCV 2022 Workshops",
year="2023",
publisher="Springer Nature Switzerland",
address="Cham",
pages="179--195",
isbn="978-3-031-25082-8"
}

@article{Shi2016,
  title={Automatic road crack detection using random structured forests},
  author={Shi, Yong and Cui, Limeng and Qi, Zhiquan and Meng, Fan and Chen, Zhensong},
  journal={IEEE Transactions on Intelligent Transportation Systems},
  volume={17},
  number={12},
  pages={3434--3445},
  year={2016},
  publisher={IEEE}
}

@InProceedings{Sudre2017,
author="Sudre, Carole H.
and Li, Wenqi
and Vercauteren, Tom
and Ourselin, Sebastien
and Jorge Cardoso, M.",
editor="Cardoso, M. Jorge
and Arbel, Tal
and Carneiro, Gustavo
and Syeda-Mahmood, Tanveer
and Tavares, Jo{\~a}o Manuel R.S.
and Moradi, Mehdi
and Bradley, Andrew
and Greenspan, Hayit
and Papa, Jo{\~a}o Paulo
and Madabhushi, Anant
and Nascimento, Jacinto C.
and Cardoso, Jaime S.
and Belagiannis, Vasileios
and Lu, Zhi",
title="Generalised Dice Overlap as a Deep Learning Loss Function for Highly Unbalanced Segmentations",
booktitle="Deep Learning in Medical Image Analysis and Multimodal Learning for Clinical Decision Support ",
year="2017",
publisher="Springer International Publishing",
address="Cham",
pages="240--248",
isbn="978-3-319-67558-9"
}

@article{Eyke2021,
  author    = {Eyke Hüllermeier and Willem Waegeman},
  title     = {Aleatoric and epistemic uncertainty in machine learning: an introduction to concepts and methods},
  journal   = {Machine Learning},
  volume    = {110},
  number    = {3},
  pages     = {457--506},
  year      = {2021},
  doi       = {10.1007/s10994-021-05946-3},
  url       = {https://doi.org/10.1007/s10994-021-05946-3},
  issn      = {1573-0565}
}

@article{Faber2005,
    author = {Faber, Michael Havbro},
    title = "{On the Treatment of Uncertainties and Probabilities in Engineering Decision Analysis}",
    journal = {Journal of Offshore Mechanics and Arctic Engineering},
    volume = {127},
    number = {3},
    pages = {243-248},
    year = {2005},
    month = {03},
    issn = {0892-7219},
    doi = {10.1115/1.1951776},
}

@InProceedings{Khanzhina2023,
    author    = {Khanzhina, Natalia and Kashirin, Maxim and Filchenkov, Andrey},
    title     = {New Bayesian Focal Loss Targeting Aleatoric Uncertainty Estimate: Pollen Image Recognition},
    booktitle = {Proceedings of the IEEE/CVF Conference on Computer Vision and Pattern Recognition (CVPR) Workshops},
    month     = {June},
    year      = {2023},
    pages     = {4253-4262}
}

@article{Audrey2021,
title = {Bayesian neural networks for uncertainty quantification in data-driven materials modeling},
journal = {Computer Methods in Applied Mechanics and Engineering},
volume = {386},
pages = {114079},
year = {2021},
issn = {0045-7825},
doi = {https://doi.org/10.1016/j.cma.2021.114079},
url = {https://www.sciencedirect.com/science/article/pii/S0045782521004102},
author = {Audrey Olivier and Michael D. Shields and Lori Graham-Brady}
}

@article{CHEN2021354,
title = {Adaptive sparse dropout: Learning the certainty and uncertainty in deep neural networks},
journal = {Neurocomputing},
volume = {450},
pages = {354-361},
year = {2021},
issn = {0925-2312},
doi = {https://doi.org/10.1016/j.neucom.2021.04.047},
author = {Yuanyuan Chen and Zhang Yi}
}

@INPROCEEDINGS{Lin2022,
  author={Lin, Shuyu and Clark, Ronald and Trigoni, Niki and Roberts, Stephen},
  booktitle={ICASSP 2022 - 2022 IEEE International Conference on Acoustics, Speech and Signal Processing (ICASSP)}, 
  title={Uncertainty Estimation with a VAE-Classifier Hybrid Model}, 
  year={2022},
  volume={},
  number={},
  pages={3548-3552},
  doi={10.1109/ICASSP43922.2022.9747520}}

@article{Xuming2022,
title = {Detecting out-of-distribution samples via variational auto-encoder with reliable uncertainty estimation},
journal = {Neural Networks},
volume = {145},
pages = {199-208},
year = {2022},
issn = {0893-6080},
doi = {https://doi.org/10.1016/j.neunet.2021.10.020},
author = {Xuming Ran and Mingkun Xu and Lingrui Mei and Qi Xu and Quanying Liu}
}

@InProceedings{chen2024,
    author    = {Chen, Zhuangzhuang and Lai, Zhuonan and Chen, Jie and Li, Jianqiang},
    title     = {Mind Marginal Non-Crack Regions: Clustering-Inspired Representation Learning for Crack Segmentation},
    booktitle = {Proceedings of the IEEE/CVF Conference on Computer Vision and Pattern Recognition (CVPR)},
    month     = {June},
    year      = {2024},
    pages     = {12698-12708}
}

@inbook{Hu2019,
author = {Hu, Xiaoling and Fuxin, Li and Samaras, Dimitris and Chen, Chao},
title = {Topology-preserving deep image segmentation},
year = {2019},
publisher = {Curran Associates Inc.},
address = {Red Hook, NY, USA},
booktitle = {Proceedings of the 33rd International Conference on Neural Information Processing Systems},
articleno = {508},
numpages = {12}
}

@INPROCEEDINGS{Seyedhosseini2013,
  author={Seyedhosseini, Mojtaba and Sajjadi, Mehdi and Tasdizen, Tolga},
  booktitle={2013 IEEE International Conference on Computer Vision}, 
  title={Image Segmentation with Cascaded Hierarchical Models and Logistic Disjunctive Normal Networks}, 
  year={2013},
  volume={},
  number={},
  pages={2168-2175},
  doi={10.1109/ICCV.2013.269}}

@misc{goo2024hybridsegmentor,
      title={Hybrid-Segmentor: A Hybrid Approach to Automated Fine-Grained Crack Segmentation in Civil Infrastructure}, 
      author={June Moh Goo and Xenios Milidonis and Alessandro Artusi and Jan Boehm and Carlo Ciliberto},
      year={2024},
      eprint={2409.02866},
      archivePrefix={arXiv},
      primaryClass={cs.CV},
      url={https://arxiv.org/abs/2409.02866}, 
}

@article{Kendall2017,
  title={What uncertainties do we need in bayesian deep learning for computer vision?},
  author={Kendall, Alex and Gal, Yarin},
  journal={Advances in neural information processing systems},
  volume={30},
  year={2017}
}

@article{Ketson2025,
  author    = {Ketson R. M. dos Santos and Adrien G. J. Chassignet and Bryan G. Pantoja-Rosero and Amir Rezaie and Onaïa J. Savary and Katrin Beyer},
  title     = {Uncertainty quantification for a deep learning models for image-based crack segmentation},
  journal   = {Journal of Civil Structural Health Monitoring},
  volume    = {15},
  number    = {4},
  pages     = {1231--1269},
  year      = {2025},
  month     = {April},
  doi       = {10.1007/s13349-024-00879-6},
  url       = {https://doi.org/10.1007/s13349-024-00879-6},
  issn      = {2190-5479},
}

@article{Rahul2023,
title = {Epistemic and aleatoric uncertainty quantification for crack detection using a Bayesian Boundary Aware Convolutional Network},
journal = {Reliability Engineering And System Safety},
volume = {240},
pages = {109547},
year = {2023},
issn = {0951-8320},
doi = {https://doi.org/10.1016/j.ress.2023.109547},
author = {Rahul Rathnakumar and Yutian Pang and Yongming Liu},
}

@InProceedings{Selvan2020,
author="Selvan, Raghavendra
and Faye, Frederik
and Middleton, Jon
and Pai, Akshay",
editor="Liu, Mingxia
and Yan, Pingkun
and Lian, Chunfeng
and Cao, Xiaohuan",
title="Uncertainty Quantification in Medical Image Segmentation with Normalizing Flows",
booktitle="Machine Learning in Medical Imaging",
year="2020",
address="Cham",
pages="80--90",
isbn="978-3-030-59861-7"
}

@inproceedings{Zhang20222,
author = {Zhang, Wei and Zhang, Xiaohong and Huang, Sheng and Lu, Yuting and Wang, Kun},
title = {A Probabilistic Model for Controlling Diversity and Accuracy of Ambiguous Medical Image Segmentation},
year = {2022},
isbn = {9781450392037},
publisher = {Association for Computing Machinery},
address = {New York, NY, USA},
url = {https://doi.org/10.1145/3503161.3548115},
doi = {10.1145/3503161.3548115},
booktitle = {Proceedings of the 30th ACM International Conference on Multimedia},
pages = {4751–4759},
numpages = {9},
location = {Lisboa, Portugal},
series = {MM '22}
}

@INPROCEEDINGS{Zbinden2023,
  author={Zbinden, Lukas and Doorenbos, Lars and Pissas, Theodoros and Huber, Adrian Thomas and Sznitman, Raphael and Márquez-Neila, Pablo},
  booktitle={2023 IEEE/CVF International Conference on Computer Vision (ICCV)}, 
  title={Stochastic Segmentation with Conditional Categorical Diffusion Models}, 
  year={2023},
  volume={},
  number={},
  pages={1119-1129},
  doi={10.1109/ICCV51070.2023.00109}}

@INPROCEEDINGS{Cheng2021,
  author={Cheng, Mingfei and Zhao, Kaili and Guo, Xuhong and Xu, Yajing and Guo, Jun},
  booktitle={2021 IEEE/CVF International Conference on Computer Vision (ICCV)}, 
  title={Joint Topology-preserving and Feature-refinement Network for Curvilinear Structure Segmentation}, 
  year={2021},
  volume={},
  number={},
  pages={7127-7136},
  doi={10.1109/ICCV48922.2021.00706}}

@article{Qin2012,
title = {CrackTree: Automatic crack detection from pavement images},
journal = {Pattern Recognition Letters},
volume = {33},
number = {3},
pages = {227-238},
year = {2012},
issn = {0167-8655},
doi = {https://doi.org/10.1016/j.patrec.2011.11.004},
url = {https://www.sciencedirect.com/science/article/pii/S0167865511003795},
author = {Qin Zou and Yu Cao and Qingquan Li and Qingzhou Mao and Song Wang},
}

@article{Perez2018, 
title={FiLM: Visual Reasoning with a General Conditioning Layer}, 
volume={32}, 
DOI={10.1609/aaai.v32i1.11671}, 
number={1}, 
journal={Proceedings of the AAAI Conference on Artificial Intelligence}, 
author={Perez, Ethan and Strub, Florian and de Vries, Harm and Dumoulin, Vincent and Courville, Aaron}, 
year={2018}, 
month={Apr.} 
}

@inproceedings{WongToi2023,
title={Understanding Pathologies of Deep Heteroskedastic Regression},
author={Eliot Wong-Toi and Alex James Boyd and Vincent Fortuin and Stephan Mandt},
booktitle={The 40th Conference on Uncertainty in Artificial Intelligence},
year={2024},
url={https://openreview.net/forum?id=n5faLvrsA0}
}

@article{Mengxue2026,
title = {Uncertainty-guided progressive learning for pavement crack segmentation under challenging visual degradation in smartphone-based imaging},
journal = {Measurement},
volume = {273},
pages = {121178},
year = {2026},
issn = {0263-2241},
doi = {https://doi.org/10.1016/j.measurement.2026.121178},
url = {https://www.sciencedirect.com/science/article/pii/S0263224126008870},
author = {Mengxue Guo and Hua Huang and Fankui Zeng and Mengzhao Nie and Mingxia Dang},
}

@article{Chen2025,
title = {SelectSeg: Uncertainty-based selective training and prediction for accurate crack segmentation under limited data and noisy annotations},
journal = {Reliability Engineering System Safety},
volume = {259},
pages = {110909},
year = {2025},
issn = {0951-8320},
doi = {https://doi.org/10.1016/j.ress.2025.110909},
url = {https://www.sciencedirect.com/science/article/pii/S0951832025001127},
author = {Chen Zhang and Mahdi Bahrami and Dhanada K. Mishra and Matthew M.F. Yuen and Yantao Yu and Jize Zhang},
}

@inproceedings{Seitzer2022,
title={On the Pitfalls of Heteroscedastic Uncertainty Estimation with Probabilistic Neural Networks},
author={Maximilian Seitzer and Arash Tavakoli and Dimitrije Antic and Georg Martius},
booktitle={International Conference on Learning Representations},
year={2022},
url={https://openreview.net/forum?id=aPOpXlnV1T}
}

@article{Jianming2025,
title = {Crack segmentation network via difference convolution-based encoder and hybrid CNN-Mamba multi-scale attention},
journal = {Pattern Recognition},
volume = {167},
pages = {111723},
year = {2025},
issn = {0031-3203},
doi = {https://doi.org/10.1016/j.patcog.2025.111723},
url = {https://www.sciencedirect.com/science/article/pii/S0031320325003838},
author = {Jianming Zhang and Shigen Zhang and Dianwen Li and Jianxin Wang and Jin Wang},
}

@article{Xiangkun2025,
title = {CCDFormer: A dual-backbone complex crack detection network with transformer},
journal = {Pattern Recognition},
volume = {161},
pages = {111251},
year = {2025},
issn = {0031-3203},
doi = {https://doi.org/10.1016/j.patcog.2024.111251},
url = {https://www.sciencedirect.com/science/article/pii/S0031320324010021},
author = {Xiangkun Hu and Hua Li and Yixiong Feng and Songrong Qian and Jian Li and Shaobo Li},
}

\end{document}